\documentclass[conference]{IEEEtran}
\IEEEoverridecommandlockouts
\usepackage{eso-pic}
\usepackage{cite}
\usepackage{amsmath,amssymb,amsfonts}
\usepackage{graphicx}
\usepackage{xcolor}
\usepackage{url}
\usepackage{textcomp}
\usepackage[normalem]{ulem}
\usepackage{newtxtext}

\renewcommand{\baselinestretch}{0.975} 

\newcommand{\sysname}{\textsc{DLaaS}}

\newcommand{\crv}[1]{\textcolor{black}{#1}}

\usepackage{booktabs}
\usepackage{listings}
\newcommand{\hl}[1]{\textcolor{demoacc}{\textbf{(#1)}}}

\usepackage{tikz}
\usetikzlibrary{positioning,arrows.meta}
\definecolor{demoacc}{HTML}{E8590C}
\tikzset{demobadge/.style={circle, fill=demoacc, text=white,
  font=\bfseries\scriptsize, inner sep=0pt, minimum size=4.2mm,
  draw=white, line width=0.5pt}}
\newcommand{\filmpanel}[2]{%
  \begin{minipage}[c][3.4cm][c]{3.4cm}\centering
    \IfFileExists{#1}{\includegraphics[width=3.25cm,height=3.25cm,keepaspectratio]{#1}}{\scriptsize\itshape #2}%
  \end{minipage}%
}

\renewcommand{\baselinestretch}{0.975}

\AtBeginDocument{%
  \setlength{\textfloatsep}{3pt plus 1pt minus 1pt}%
  \setlength{\dbltextfloatsep}{3pt plus 1pt minus 1pt}%
  \setlength{\floatsep}{3pt plus 1pt minus 1pt}%
  \setlength{\dblfloatsep}{3pt plus 1pt minus 1pt}%
  \setlength{\intextsep}{3pt plus 1pt minus 1pt}%
  \setlength{\abovecaptionskip}{2pt}%
  \setlength{\belowcaptionskip}{1pt}%
}
\newcommand{\placetextbox}[3]{%
  \AddToShipoutPictureFG*{%
    \put(\LenToUnit{#1\paperwidth},\LenToUnit{#2\paperheight}){%
      \vtop{{\null}\makebox[0pt][c]{#3}}}%
  }%
}

\placetextbox{.5}{0.034}{%
\fbox{\parbox{0.82\paperwidth}{%
\centering\fontsize{6.2}{7}\selectfont
\textcopyright~IFIP, 2026. This is the author's version of a work accepted for publication in the 22nd International Conference on Network and Service Management (CNSM 2026).
}}}
\begin{document}

\title{Distributed Learning as a Service:\\
The Developer's Perspective}

\author{\IEEEauthorblockN{Tianyue Chu$^{1}$, Filippo Vannella$^{1}$, Dimitra Tsigkari$^{1}$, Paula Delgado-Santos$^{1}$ \\
Fernando López$^{1}$, Pablo Gomez Guerrero$^{1}$, Sotirios Spantideas$^{2}$, David Solans Noguero$^{1}$}
\IEEEauthorblockA{$^{1}$\textit{Telefónica Scientific Research}, \textit{Telefónica}, Madrid, Spain \\
$^{2}$\textit{National and Kapodistrian University of Athens}, Psachna, Evia, Greece \\
}

}

\maketitle
\begin{abstract}
Application developers of distributed learning services face challenges that a typical federated learning loop does not address. Specifically, the model updates can still leak private data, devices might not be able to participate in the training due to limited resources, a single aggregator might not be able to scale, and the transmissions of model weights induce a considerable bandwidth cost. This paper demonstrates \sysname{} (Distributed Learning as a Service) from the developer's vantage point. Using a single admin dashboard, the developer initiates a distributed/federated learning job and is able to activate Differential Privacy (DP), Split Learning (SL), Hierarchical Aggregation (HA), and Knowledge Distillation (KD) as declarative options, with no change to the clients' code. We demonstrate the complete service lifecycle on an industrial smart-home Wake-up Word (WuW) task, using the ``Ok Aura'' dataset. Once the developer initiates a distributed learning job by toggling DP, SL, HA, and KD in the admin dashboard, the system dispatches the job to a set of Android clients and Dockerized helper aggregators. In the demonstration, these mechanisms run live across configurations. Then, the clients train the model locally and return their updates. The trained model is served to a consumer-side Android application that performs on-device WuW detection on a live microphone stream. In particular, the conference attendees will be invited to speak the trigger phrase and monitor in real time the per-class confidence and inference latency.  Finally, we release the source code and short video walkthroughs of these configurations.
\end{abstract}

\section{Introduction}
Machine Learning (ML) is increasingly trained on Internet of Things (IoT) and other edge devices, close to where data is produced. Such data often cannot be collected in one place, because it is private or costly to move.
A common response is distributed learning, which spreads the training across many devices rather than one server. Federated Learning (FL) is a widely used method of distributed learning: each device keeps its own data and shares only model updates~\cite{mcmahan2017fedavg}. 
Federated Learning as a Service (FLaaS)~\cite{kourtellis2020flaas,katevas2022flaas} is a service that allows an application developer to manage the FL training without orchestrating the underlying system. 
Deploying FL as a production service, however, raises four challenges a basic FL loop does not address. First, the exchanged updates may still reveal private information. Second, devices with limited memory may be unable to train large models. Third, a single aggregator limits scalability as more clients join. Fourth, transmitting a large model in every round is costly over constrained links. In a real deployment, such as a smart home, these challenges arise together.

\sysname{} (Distributed Learning as a Service) overcomes these obstacles by extending the open-source FLaaS framework~\cite{kourtellis2020flaas,katevas2022flaas} with four mechanisms: Differential Privacy (DP), Split Learning (SL), Hierarchical Aggregation (HA), and Knowledge Distillation (KD). Each of these mechanisms appears as a declarative switch in the same job specification rather than a separate system to integrate. 
To the best of our knowledge, \sysname{} is the first framework
to jointly support all four mechanisms while preserving access
to the FL functionalities of FLaaS.
This demo presents \sysname{} from the developer's perspective, showing what it is  like to configure, launch, and watch a distributed learning job running from end to end. 

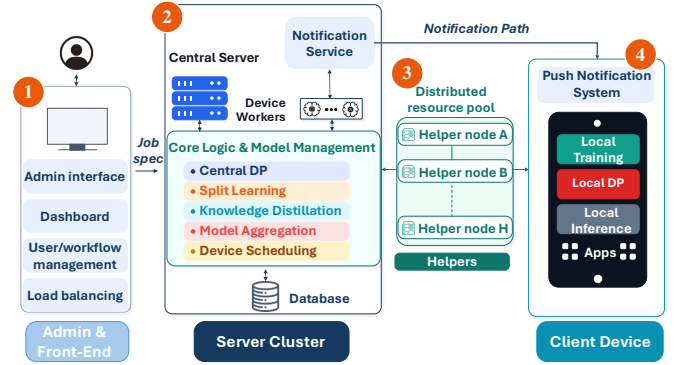
\begin{figure}[t]
\centering
\IfFileExists{Figures/DLaaS_Architecture.pdf}{%
  \begin{tikzpicture}
    \node[anchor=south west, inner sep=0] (img) at (0,0)
      {\includegraphics[width=\columnwidth]{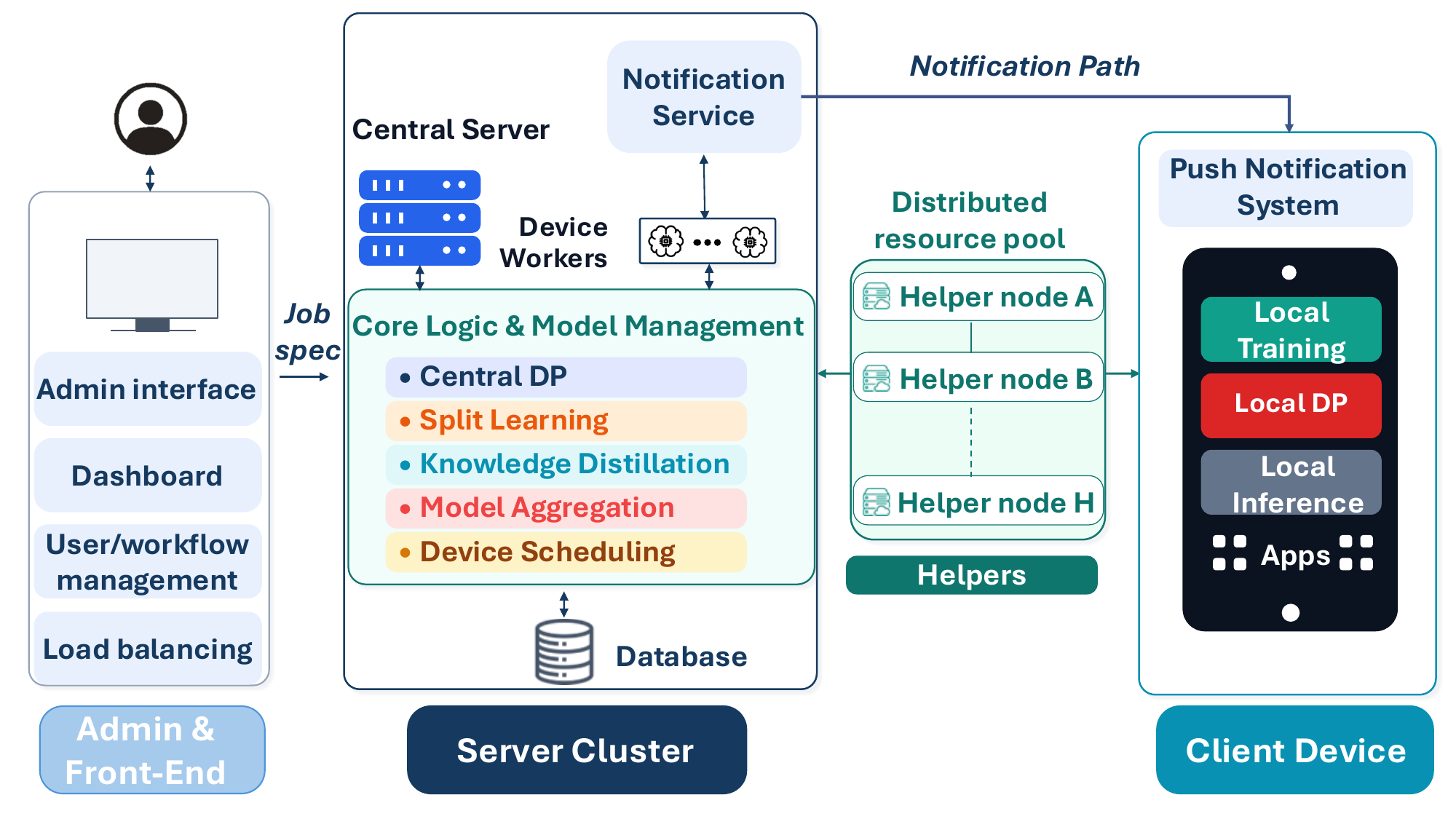}};
    \begin{scope}[x={(img.south east)}, y={(img.north west)}]
      \node[demobadge] at (0.032,0.75) {1}; 
      \node[demobadge] at (0.24,0.95) {2}; 
      \node[demobadge] at (0.6,0.8) {3}; 
      \node[demobadge] at (0.95,0.85) {4}; 
    \end{scope}
  \end{tikzpicture}%
}{%
  \setlength{\fboxsep}{8pt}%
  \fbox{\parbox[c][3.0cm][c]{0.88\columnwidth}{\centering\itshape [Architecture figure not found at \texttt{Figures/DLaaS\_Architecture.pdf}.]}}%
}
\caption{\textbf{\sysname{} architecture} annotated for the demonstration as a job flows through the system: \textbf{(1)} the developer composes the job at the admin front-end (DP mechanism and its $(\varepsilon,\delta)$ values, the SL, HA, and KD toggles) and submits the specifications; \textbf{(2)} the server cluster compiles the given specs into a per-round plan and runs the server-side core logic (central DP, SL, KD, model aggregation, and device scheduling), driving the distributed learning round loop; \textbf{(3)} the helper tier performs partial aggregation over client clusters and can host the SL server-side part; and \textbf{(4)} each client device runs local training, local DP, and sends the trained model back to the server.}
\label{fig:arch}
\end{figure}

We demonstrate \sysname{} on an industrial smart-home Wake-up Word (WuW) detection workload, the ``Ok Aura'' task~\cite{lopez2024okey}, in which an always-on device must recognize a trigger phrase on-device. 
\begin{figure*}[t]
\centering
\includegraphics[width=\textwidth]{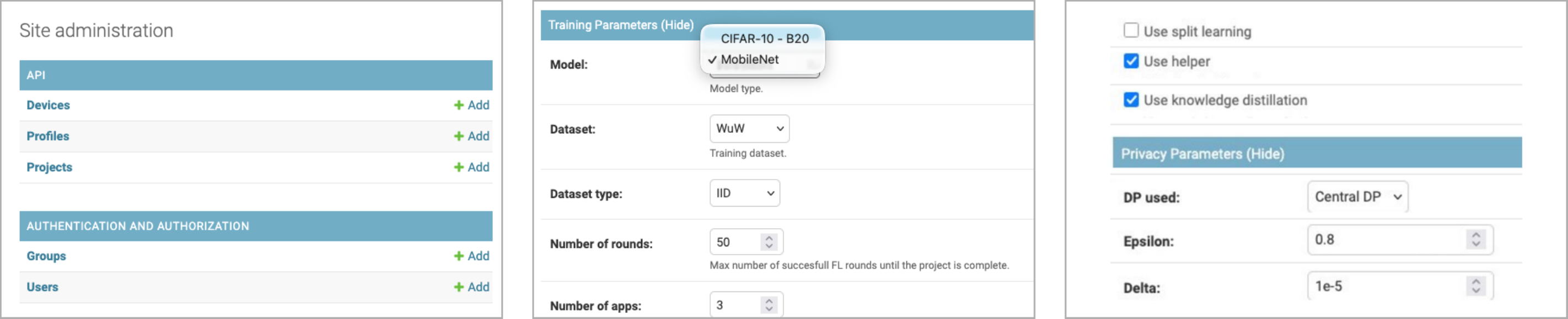}\\[2pt]
\makebox[0.333\textwidth]{\footnotesize (a) create and manage projects}%
\makebox[0.333\textwidth]{\footnotesize (b) select model, dataset, and FL rounds}%
\makebox[0.333\textwidth]{\footnotesize (c) configure privacy and SL\,/\,HA\,/\,KD modules}
\caption{\textbf{(1) Compose} --- the server-side admin interface (Django admin). (a) the admin home, where projects are created, scheduled, and managed; (b) a project's training configuration, selecting the model, dataset, and number of FL rounds; and (c) the DP mechanism with its $(\varepsilon,\delta)$ configuration and the SL, HA, and KD module toggles. The whole job is composed from this one interface.}
\label{fig:f_compose}
\end{figure*}
The WuW detection task is a representative edge-learning workload since it combines privacy-sensitive audio data, continuously operating resource-constrained devices, and large-scale deployment requirements.
The demonstration shows how privacy, device-awareness, aggregation scale, and communication cost can be turned on and off as first-class service policies. To make the walkthrough reproducible beyond the demo, we also publish the source code and a short video extract for each different configuration.
\section{\sysname{} Overview}
Fig.~\ref{fig:arch} presents the \sysname{} architecture. It has three tiers: (i) a central coordinator \crv{(admin front-end~(1) and server cluster~(2) in the figure)}, (ii) a helper tier~\crv{(3)}, and (iii) an on-device client SDK~\crv{(4)}.
The coordinator, a Django REST service, exposes the control plane and holds the federated state. The helper tier is a set of stateless Dockerized FastAPI aggregators that perform partial aggregation. The client SDK runs local training and inference on Android. Through the control plane, a job is declared rather than coded. The developer selects the dataset and training setup, a DP mechanism with its $(\varepsilon,\delta)$ budget, FL or SL execution, a flat or hierarchical topology, and toggles distillation. Given these choices, the coordinator compiles the specification into a per-round plan and pushes it to the helpers and clients. Depending on which modules are active, the data plane then carries the model weights, the intermediate activations~(under SL), or the distilled updates.

We now elaborate on the four modules of \sysname{}.
\textit{(i)~DP:}~\sysname{} supports central and local DP. The central DP clips per-client updates and adds calibrated Gaussian noise at the server, with the noise multiplier derived offline from the target $(\varepsilon,\delta)$ under R\'enyi-DP composition \cite{mironov2017renyi} using the Opacus subsampled-Gaussian accountant~\cite{opacus}. Local DP privatizes each client's update on-device via the analytic Gaussian mechanism~\cite{dwork2014algorithmic}, removing the trusted-aggregator assumption at the cost of higher noise. \textit{(ii) SL:} for devices with limited memory, \sysname{} splits the model at a configurable cut layer, so the client computes the \crv{frozen} first part and streams activations to a server (or helper) that \crv{trains the remaining layers}~\cite{thapa2022splitfed}. \textit{(iii) HA:} a helper tier performs weighted partial aggregation over client clusters and forwards a single result upstream, reducing server-side fan-in from $O(N)$ to $O(H)$ while preserving FedAvg equivalence~\cite{liu2020hierfavg}; helpers can additionally host the SL server-side part, so SL and HA compose. \textit{(iv) KD:} \sysname{} distills a high-capacity teacher into a compact student server-side and ships only the student to clients \cite{hinton2015distilling}, reducing per-round model transmission and on-device footprint while leaving the client training loop unchanged. \crv{All four are exposed as independent fields of the job specification and can be enabled together in one job.}

\section{Demonstration} \label{sec:demonstration}
\begin{figure}[t]
\centering
\begin{lstlisting}[basicstyle=\fontsize{6.8}{7.3}\selectfont\ttfamily,frame=single,escapechar=|,xleftmargin=2pt,framexleftmargin=2pt,columns=fullflexible,keepspaces=true,aboveskip=0pt,belowskip=0pt]
$ python manage.py tick        (project 87, round 0)
|\hl{a}| Reported models are enough (Ratio 1.00).
    Round '0' is complete. Creating next round: 1
|\hl{b}| Using helper. Submitting group 1/1 to helper
    on port 8500  [NET] uplink_bytes=8000259
    downlink_bytes=8000189 latency_s=0.579
    [DEBUG] Helper container logs:
      INFO: Uvicorn running on http://0.0.0.0:8500
|\hl{c}|   INFO: "POST /aggregate HTTP/1.1" 200 OK
    Helper 1 (port 8500, size 1) finished in 3.72 s
    Helper-based aggregation complete.
    Model weights saved to projects/87/1/model_weights.bin
|\hl{d}| Sending train request: users:['test_user1']
    data: {'type': 'train', 'project': 87, 'round': 1,
      'trainingMode': 'BASELINE', 'localDP': 0, ...}
    Push sent successfully.
\end{lstlisting}
\caption{\textbf{(2) Dispatch \& aggregate} --- \crv{excerpt of the system coordinator terminal, with the relevant messages marked: \hl{a}~round~0 is complete; \hl{b}~the round is dispatched to a helper container; \hl{c}~the helper aggregates and the model is saved; \hl{d}~round~1 is pushed to the clients.}}
\label{fig:f_dispatch}
\end{figure}
\begin{figure}[t]
\centering
\includegraphics[height=2.8cm]{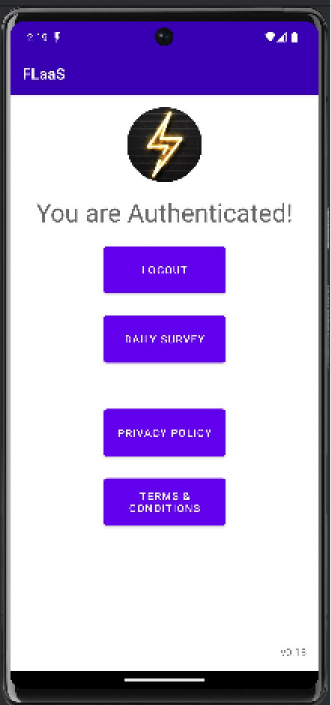}\hfill
\includegraphics[height=2.8cm]{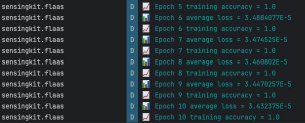}
\caption{\textbf{(3) Local training} --- the Android client. \crv{Left: the authenticated app on the emulator. Right: the Logcat lines reporting per-epoch training loss and accuracy (timestamp, PID, and tag columns omitted for readability).}}
\label{fig:f_train}
\end{figure}

\begin{figure}[t]
\centering
\includegraphics[height=6.3cm,keepaspectratio]{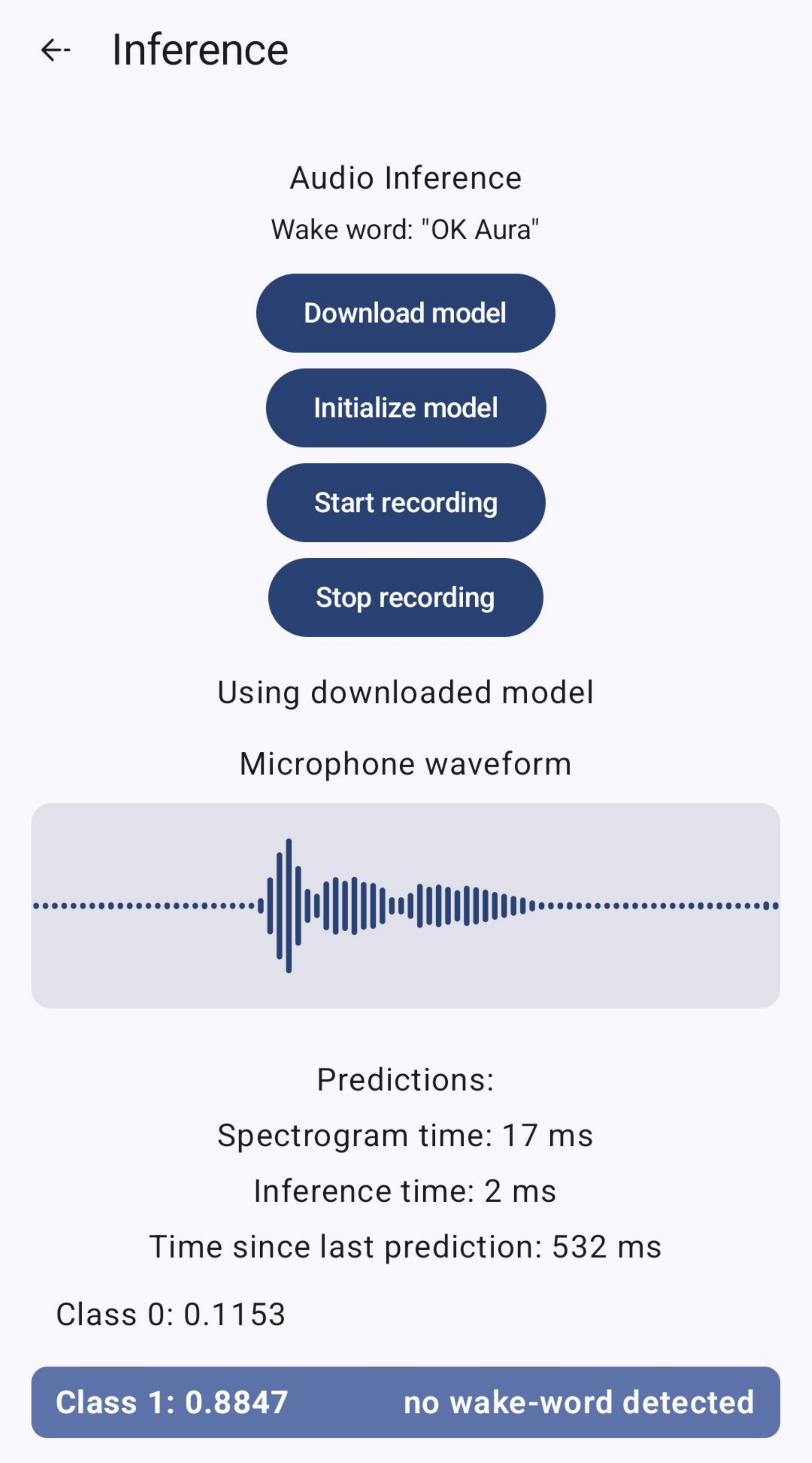}
\caption{\textbf{(4) Detect} --- the consumer app. It downloads the trained global model and reports live on the ``Ok Aura'' detection task with per-class confidence and inference latency. }
\label{fig:f_detect}
\end{figure}

 
The demonstration walks developers through the full \sysname{} service loop on the WuW workload, with the live screen for each step shown in Figs.~\ref{fig:f_compose}--\ref{fig:f_detect}. A job is composed once (step~1); the system coordinator and clients then repeat steps~2 and~3 for $T$ federated rounds before the consumer app downloads the trained global model and runs live ``Ok Aura'' detection (step~4). 
\crv{In the demo, one complete FL round (local training followed by global aggregation) runs live.
Reaching a converged model takes many such rounds, so the model served to the consumer app in step~4 is taken from a full training run performed beforehand.} 


\textit{(1) Compose.} From the server-side admin interface, the developer schedules a project, selecting the model, dataset, and number of FL rounds, and configures privacy and modules: the DP mechanism (central or local) and budget $\varepsilon$, and the SL, HA (with $H$ helpers), and KD toggles (Fig.~\ref{fig:f_compose}).

\textit{(2) Dispatch and aggregate.} The coordinator compiles the spec into a fixed per-round plan and drives the round loop: it recruits clients, dispatches the round, and the helper tier aggregates the returned updates into the new global model before pushing the next round (Fig.~\ref{fig:f_dispatch}). Toggling HA on the same project makes its effect directly observable at the coordinator: the server then receives one aggregated result per helper rather than every client update, so the per-round fan-in drops from the client count to the number of helpers $H$.

\textit{(3) Local training.} Each client runs the FL round on an Android device (a Pixel profile on an \texttt{arm64-v8a} emulator), training the model head on its local samples and returning the update; the Logcat reports the per-epoch loss and accuracy (Fig.~\ref{fig:f_train}). Under SL it instead exports bottleneck activations, and under local DP it adds noise to the update on-device.

\textit{(4) Detect.} After the rounds complete, the trained global model is served to a consumer-side Android app that runs continuous on-device detection; an attendee can say \textit{``Ok Aura''} and see the per-class confidence and an inference latency of only a few milliseconds~(Fig.~\ref{fig:f_detect}), closing the loop from configuration to a working consumer experience.

\textit{Evaluation of \sysname{}.} Since a few live rounds cannot fully showcase  the performance of \sysname{},
\crv{the quantitative trade-offs below come from our offline evaluation on the same testbed, on CIFAR-10 and on the WuW task. There, we observe that}
SL trims peak on-device memory by 33\%; KD shrinks the broadcast model by 88.8\% on CIFAR-10 and 92.4\% on the WuW task, a 78.5\,MB teacher distilled to a 5.9\,MB student; and HA replaces the server's $O(N)$ per-round fan-in with $O(H)$, one helper projected to serve roughly 139 clients within an 8\,GB budget. To surface these effects, especially the SL and KD savings, the demonstration runs the heavier CIFAR-10 model.
\section{Demo Setup and Usage}
The demonstration runs self-contained on a single laptop that hosts the coordinator, the Dockerized helper aggregators, and the emulated Android client, together with one Android phone running the WuW inference application. Network connectivity is required, since the coordinator sends round notifications to the client devices.
The code is open source,\footnote{\url{https://github.com/Telefonica-Scientific-Research/DLaaS-Server}} and we provide videos of a complete run for those configurations:
\begin{itemize}
    \item FL training: \url{https://youtu.be/zvImEQucwX8}
    \item DP: \url{https://youtu.be/H4SA9UqQjOM}
    \item SL: \url{https://youtu.be/6CqTziGy3J4}
    \item HA: \url{https://youtu.be/XeBnISOqaSw}
    \item KD: \url{https://youtu.be/XiZS16lx-2s}
    \item WuW training: \url{https://youtu.be/ppHJI2smsC8} 
    \item WuW inference: \url{https://youtu.be/QqXN84qxETo}
\end{itemize}

\section{Conclusion}
This demonstration puts a developer in control of \sysname{} and shows that four capabilities usually added through separate tooling, privacy, device-awareness, scalability, and communication efficiency, can instead be exposed as composable service-level policies within a distributed learning platform. By transforming advanced distributed learning techniques into configurable operational primitives, \sysname{}  paves the way towards production-grade distributed learning systems.

\section*{Acknowledgments}
This research is supported by the European Union’s Horizon Europe research and innovation actions under grant agreement No 101168560 (CoEvolution), the European Union under TaRDIS (GA 101093006), and the Horizon MSCA Postdoctoral Fellowship OPALS (grant agreement 101210495). Views and opinions expressed are however those of the authors only and do not necessarily reflect those of the European Union. Neither the European Union nor the granting authority can be held responsible for them. 

\centering \includegraphics[width=2.87cm,trim={0cm 0.53cm 0cm 0cm},clip]{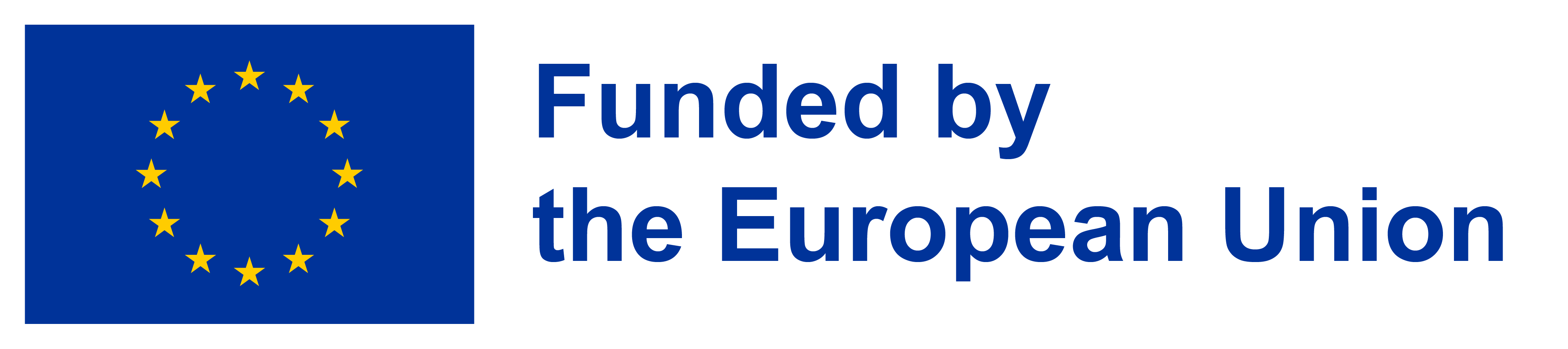}

\bibliographystyle{IEEEtran}
\bibliography{refs}

@inproceedings{kourtellis2020flaas,
  author    = {Nicolas Kourtellis and Kleomenis Katevas and Diego Perino},
  title     = {{FLaaS}: Federated Learning as a Service},
  booktitle = {Proc.\ 1st Workshop on Distributed Machine Learning
               (DistributedML)},
  year      = {2020},
  note      = {arXiv:2011.09359}
}

@inproceedings{katevas2022flaas,
  author    = {Kleomenis Katevas and Diego Perino and Nicolas Kourtellis},
  title     = {{FLaaS} -- Enabling Practical Federated Learning on Mobile
               Environments},
  booktitle = {Proc.\ 20th Annual Int.\ Conf.\ on Mobile Systems,
               Applications and Services (MobiSys, Demo)},
  pages     = {605--606},
  year      = {2022},
  publisher = {ACM},
  doi       = {10.1145/3498361.3539693}
}

@misc{opacus,
  author       = {Ashkan Yousefpour and Igor Shilov and Alexandre Sablayrolles
                  and Davide Testuggine and Karthik Prasad and Mani Malek
                  and John Nguyen and Sayan Ghosh and Akash Bharadwaj and
                  Jessica Zhao and Graham Cormode and Ilya Mironov},
  title        = {{Opacus}: User-Friendly Differential Privacy Library in
                  {PyTorch}},
  howpublished = {arXiv preprint arXiv:2109.12298},
  year         = {2021}
}

@inproceedings{mironov2017renyi,
  title={R{\'e}nyi differential privacy},
  author={Mironov, Ilya},
  booktitle={2017 IEEE 30th computer security foundations symposium (CSF)},
  pages={263--275},
  year={2017},
  organization={IEEE}
}

@inproceedings{liu2020hierfavg,
  author    = {Lumin Liu and Jun Zhang and S. H. Song and Khaled B. Letaief},
  title     = {Client-Edge-Cloud Hierarchical Federated Learning},
  booktitle = {Proc.\ of IEEE ICC},
  pages={1--6},
  year      = {2020}
}

@inproceedings{thapa2022splitfed,
  title={Splitfed: When federated learning meets split learning},
  author={Thapa, Chandra and Arachchige, Pathum Chamikara Mahawaga and Camtepe, Seyit and Sun, Lichao},
  booktitle={Proceedings of AAAI},
  volume={36},
  number={8},
  pages={8485--8493},
  year={2022}
}

@inproceedings{hinton2015distilling,
  title={Distilling the knowledge in a neural network},
  author={Hinton, Geoffrey and Vinyals, Oriol and Dean, Jeff},
  booktitle={NeurIPS Deep Learning and Representation Learning Workshop},
  year={2015}
}

@inproceedings{mcmahan2017fedavg,
  author    = {H. Brendan McMahan and Eider Moore and Daniel Ramage and Seth
               Hampson and Blaise Ag{\"u}era y Arcas},
  title     = {Communication-Efficient Learning of Deep Networks from
               Decentralized Data},
  booktitle = {Proc. of
               AISTATS},
  year      = {2017}
}

@article{dwork2014algorithmic,
  title={The algorithmic foundations of differential privacy},
  author={Dwork, Cynthia and Roth, Aaron},
  journal={Foundations and trends{\textregistered} in theoretical computer science},
  volume={9},
  number={3-4},
  pages={211--487},
  year={2014},
  publisher={Emerald Publishing Limited}
}

@misc{lopez2024okey,
  author       = {López Gavilánez, Fernando and
                  Luque Serrano, Jordi and
                  Pablo, Gómez Guerrero},
  title        = {Okey Aura Wake-up Word Dataset (Test WUWDC 2024)},
  month        = aug,
  year         = 2024,
  publisher    = {Zenodo},
  doi          = {10.5281/zenodo.13601867},
  url          = {https://doi.org/10.5281/zenodo.13601867},
}

\end{document}